%% file: wacv.tex
\documentclass[10pt,twocolumn,letterpaper]{article}
\input{math_commands.tex}

\usepackage{wacv}
\usepackage{times}
\usepackage{epsfig}
\usepackage{graphicx}
\usepackage{amsmath}
\usepackage{amssymb}

\usepackage{url}            
\usepackage{booktabs}       
\usepackage{amsfonts}       
\usepackage{color,colortbl}
\usepackage{tabularx}
\usepackage{booktabs}
\usepackage{multirow}
\usepackage{array}
\usepackage{xspace}
\usepackage{balance}
\usepackage{enumitem}
\usepackage{mathtools}
\usepackage{rotating}
\usepackage{tabulary}
\usepackage{subcaption}
\usepackage[export]{adjustbox}
\usepackage{ctable}
\usepackage{diagbox}
\usepackage{bbm}
\usepackage{makecell}
\usepackage[linesnumbered,ruled]{algorithm2e}
\usepackage[title]{appendix}

\graphicspath{{./figs/}}

\def\wacvPaperID{996} 

\wacvfinalcopy 

\ifwacvfinal
\def\assignedStartPage{1} 
\fi

\ifwacvfinal
\usepackage[breaklinks=true,bookmarks=false]{hyperref}
\else
\usepackage[pagebackref=true,breaklinks=true,colorlinks,bookmarks=false]{hyperref}
\fi

\ifwacvfinal
\else
\fi

\begin{document}

\title{Improving Tail-Class Representation with Centroid Contrastive Learning}

\author{Anthony Meng Huat Tiong$^{1,2}$ \qquad Junnan Li$^{1}$ \qquad
Guosheng Lin$^{2}$ \qquad Boyang Li$^{2}$ \\
\qquad Caiming Xiong$^{1}$ \qquad
Steven C.H. Hoi$^{1}$\\
\and
$^{1}$Salesforce Research \qquad $^{2}$Nanyang Technological University\\
{\tt\small  \{anthony.tiong,junnan.li,cxiong,shoi\}@salesforce.com} \qquad
{\tt\small  \{gslin,boyang.li\}@ntu.edu.sg}}
\maketitle
\input{sec_abstract}
\input{sec_introduction}

\input{sec_literature}

\input{sec_method}

\input{sec_experiment}

\input{sec_conclusion}
\input{sec_acknowledgment}

{\small
\bibliographystyle{ieee_fullname}
\bibliography{bib}
}

\end{document}

%% file: math_commands.tex
\usepackage{amsmath,amsfonts,bm}

\newcommand{\Vect}[1]{\boldsymbol{#1}}

\def\eqref#1{equation~\ref{#1}}

\def\1{\bm{1}}

\DeclareMathAlphabet{\mathsfit}{\encodingdefault}{\sfdefault}{m}{sl}
\SetMathAlphabet{\mathsfit}{bold}{\encodingdefault}{\sfdefault}{bx}{n}



%% file: sec_abstract.tex
\begin{abstract}
In vision domain, large-scale natural datasets typically exhibit long-tailed distribution which has large class imbalance between head and tail classes. 
This distribution poses difficulty in learning good representations for tail classes. 
Recent developments have shown good long-tailed model can be learnt by decoupling the training into representation learning and classifier balancing. 
However, these works pay insufficient consideration on the long-tailed effect on representation learning.
In this work, we propose interpolative centroid contrastive learning (ICCL) to improve long-tailed representation learning. 
ICCL interpolates two images from a class-agnostic sampler and a class-aware sampler, and trains the model such that the representation of the interpolative image can be used to retrieve the centroids for both source classes.
We demonstrate the effectiveness of our approach on multiple long-tailed image classification benchmarks. 
Our result shows a significant accuracy gain of $2.8$\% on the iNaturalist 2018 dataset with a real-world long-tailed distribution.
\end{abstract}	

%% file: sec_introduction.tex
\section{Introduction}
\label{sec:introduction}
In recent years, deep learning algorithms have achieved impressive results in various computer vision tasks~\cite{ImageNet, coco}.
However, long-tailed recognition remains as one of the major challenges.
Different from most human-curated datasets where object classes have a balanced number of samples,
the distribution of objects in real-world is a function of Zipf's law~\cite{zipf} where a large number of tail classes have few samples.
Thus, models typically suffer a decrease in accuracy on the tail classes.
Since it is resource-intensive to curate more samples for all tail classes, it is imperative to address the challenge of long-tailed recognition.

In the literature of long-tailed recognition, typical approaches address the class imbalance issue by either data re-sampling~\cite{chawla2002smote, undersamplingbeatover, shen2016relay} or loss re-weighting techniques~\cite{khan2017cost, cbloss, shu2019meta, ldam, focal}. Re-sampling facilitates the learning of tail classes by shifting the skewed training data distribution towards the tail through undersampling or oversampling. Re-weighting modifies the loss function to encourage larger gradient contribution or decision margin of tail classes.

\input{table/fig_key_idea}
In order to correctly classify tail-class samples,
it is crucial to learn discriminative representations.
However, recent developments~\cite{ldam, decouple-longtail, bbn, cui2018large} have discovered that conventional re-sampling and re-weighting methods can lead to a suboptimal long-tailed representation learning. 
In light of these findings, various approaches proposed to decouple representation learning and classifier balancing~\cite{decouple-longtail, ldam, bbn}.
However,
most of these works focus on addressing classifier imbalance and pay inadequate attention on the negative effect of long-tailed distribution on representation learning. Thus, an intriguing question remains: can we improve long-tailed representation learning? 

We propose interpolative contrastive centroid learning (ICCL), a framework to learn more discriminative representations for tail classes.
The intuition behind our method is to use the head classes to facilitate representation learning of the tail classes.
Inspired by Mixup~\cite{mixup},
we create virtual training samples by interpolating two images from two samplers: a class-agnostic sampler which returns all images with equal probability, and a class-aware sampler which focuses more on tail-class images. 
The key idea of ICCL is shown in Figure~\ref{fig:key_idea}. 
The representation of an interpolative sample should contain information that can be used to retrieve both 
the head-class centroid and the tail-class centroid.
Specifically, we project images into a low-dimensional embedding space,
and create class centroids as average embeddings.
Given the embedding of the interpolative sample,
we query the class centroids with a contrastive similarity matching,
and train our model such that the embedding has higher similarities with the correct class centroids.

By injecting class-balanced knowledge in the form of centroids,
the proposed interpolative centroid contrastive loss 
encourages the tail centroids to be positioned discriminatively relative to the head classes. 
However, the interpolative sample may conflate a head class and a tail class. 
Thus, we adopt the regular classification loss (top branch in Figure~\ref{fig:framework}) 
to perform representation learning for head classes, so that the head classes themselves are well positioned. 
These loss components reinforce one another. 

We evaluate the effectiveness of our approach on multiple standard benchmarks of different scale, including CIFAR-LT~\cite{ldam}, ImageNet-LT~\cite{oltr} and iNaturalist 2018~\cite{inat18}. 
We compare our methods with existing state-of-the-arts including Decouple~\cite{decouple-longtail}, BBN~\cite{bbn}, and  De-confound-TDE~\cite{tang2020longtailed}.

We summarise our contribution as follows:
\vspace{-\topsep}
\begin{itemize}[leftmargin=*]
	\setlength\itemsep{0pt}
	\item We introduce interpolative centroid contrastive learning for discriminative long-tailed representation learning.
	ICCL reduces intra-class variance and increases inter-class variance by optimising the distance between sample embeddings and the class centroids.
	\item Different from previous findings that suggest class-agnostic training leads to high quality long-tailed representations,
	ICCL improves tail-class representations by addressing class imbalance with class-aware sample interpolation. 
	\item ICCL achieves significantly high performance on multiple long-tailed recognition benchmarks. 
	Notably, it achieves a substantial accuracy improvement of $2.8$\% on the large-scale iNaturalist 2018.
	We also perform ablation study to verify the effectiveness of each proposed component.
\end{itemize}
\vspace{-\topsep}

%% file: table/fig_key_idea.tex
\begin{figure}[t]
	\begin{center}
	\includegraphics[width=1\linewidth]{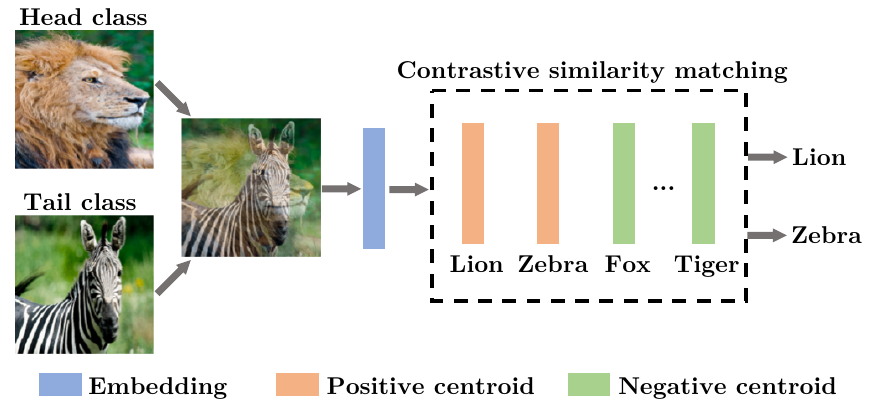}
	\vspace{-6ex}
	\end{center}
	\caption
	{ \small 
		Illustration of interpolative centroid contrastive learning. Given an
		interpolative sample, we encourage its embedding to have high 
		similarity with its corresponding head-class centroid and tail-class centroid, by contrasting it with the 
		negative centroids.
	}
	\label{fig:key_idea}
	  \vspace{-2ex}
\end{figure}

%% file: sec_literature.tex
\section{Related work}
\label{sec:literature}

\textbf{Re-sampling.} 
Re-sampling aims to address the imbalance issue from the data level.
Two main re-sampling approaches include oversampling and undersampling. 
Oversampling~\cite{chawla2002smote, chawla2003smoteboost, han2005borderline} increases the number of tail-class samples at the risk of overfitting the model, whereas undersampling might reduce the head classes diversity by decreasing their sample numbers~\cite{undersamplingbeatover, liu2008exploratory}. 
Class-balanced sampling assigns equal sampling probability for all classes, and then selects their respective images uniformly~\cite{shen2016relay}. 

\textbf{Re-weighting.} 
Re-weighting methods modify the loss function algorithmically to improve the learning of tail classes~\cite{zhou2005training, khan2017cost, jamal2020rethinking, focal, dong2017class, shu2019meta}.
Cui \etal~\cite{cbloss} introduced class-balanced loss based on the effective class samples, which improves upon the approaches that assign class weight inversely proportional to their sample number~\cite{mikolov2013distributed, huang2016learning, mahajan2018exploring}. 
Another branch of work focuses on improving the decision boundary by enforcing margin between classes~\cite{ldam, liu2016large, liu2017sphereface, wang2018cosface, deng2019arcface}.
Menon \etal~\cite{menon2020long} proposed logit adjustment to softmax loss that considers the pairwise class relative margin from statistical perspective.

\textbf{Feature Transfer.} 
Another research direction for long-tailed recognition focuses on transferring the feature representation from head to tail classes~\cite{zhong2019unequal, liu2020deep, modeltail, cui2018large, featspaceaug}. 
OLTR~\cite{oltr} transfers features of head classes to tail classes using memory centroid features. 
Rather than just having a centroid for each category, Zhu \etal~\cite{zhu2020inflated} extended OLTR by storing multiple features in both local and global memory bank to address the high intra-class variance issue.

\textbf{Decoupled Strategy.} 
Several studies have observed that applying re-sampling and re-weighting methods at the beginning of the training process 
will harm the learning of long-tailed representations~\cite{ldam, bbn, decouple-longtail}.  
BBN~\cite{bbn} introduces an extra re-balancing branch which focuses on tail classes.
The conventional uniform branch and the re-balancing branch are trained simultaneously in a curriculum learning manner.
Kang \etal~\cite{decouple-longtail} discovered that the long-tailed distribution has more negative impact on the classifier than the representation. 
They proposed a decoupled training approach, where it first performs representation learning without any re-sampling,
and then rebalances the classifier in the second stage.
Our method follows a similar two-stage training.
However, we show that long-tailed representation learning can be further improved by our proposed interpolative centroid contrastive learning even before the classifier rebalancing stage.

It is worth noting that ensemble methods have also been proposed to improve long-tailed classification accuracy~\cite{xiang2020learning, wang2020long, bbn}. Wang \etal~\cite{wang2020long} reduced the computation complexity of ensemble by only assigning uncertain samples to extra experts dynamically.

\input{table/fig_framework}
\textbf{Contrastive learning.}
Recently, contrastive learning approaches have demonstrated strong performance in self-supervised representation learning~\cite{simclr, moco, mocov2, byol, Instance, yang2020rethinking}. 
Self-supervised contrastive learning projects images into low-dimensional embeddings,
and performs similarity matching between images with different augmentations.
Two augmented images from the same source image are encouraged to have higher similarity in contrast to others.
Prototypical contrastive learning~\cite{PCL} introduces cluster-based prototypes and encourages embeddings to gather around their corresponding prototypes. 
MoPro~\cite{li2020mopro} extends this idea to weakly-supervised learning and calculates prototypes based on exponential moving average. 

Different from existing contrastive learning methods,
our ICCL operates on an interpolative sample consisting information of both classes returned by the class-agnostic and class-aware sampler.
Our contrastive loss seeks to learn a representation for the interpolative sample, such that it can be used to retrieve the centroids for both source classes.
The centroid retrieval is performed via non-parametric contrastive similarity matching in the low-dimensional space,
thus it is different from Mixup~\cite{mixup} which operates on the parametric classifier.

\textbf{Mixup.} 
As an effective data augmentation technique, Mixup~\cite{mixup} regularises the neural network by performing convex combination of training samples. Chou \etal~\cite{chou2020remix} proposed larger mixing weight for tail classes to push the decision boundary towards the head classes. 
Several works have applied Mixup to improve contrastive learning representation in \textit{self-supervised} setting~\cite{NEURIPS2020_f7cade80, kim2020mixco, shen2020rethinking,  zhou2020C2L, verma2020towards, lee2021imix}. 
In contrast, our interpolative centroid contrastive loss is a new loss designed to 
improve \textit{supervised} representation learning under long-tailed distribution.

%% file: table/fig_framework.tex
\begin{figure*}[!t]

	\centering
	\includegraphics[width=\textwidth]{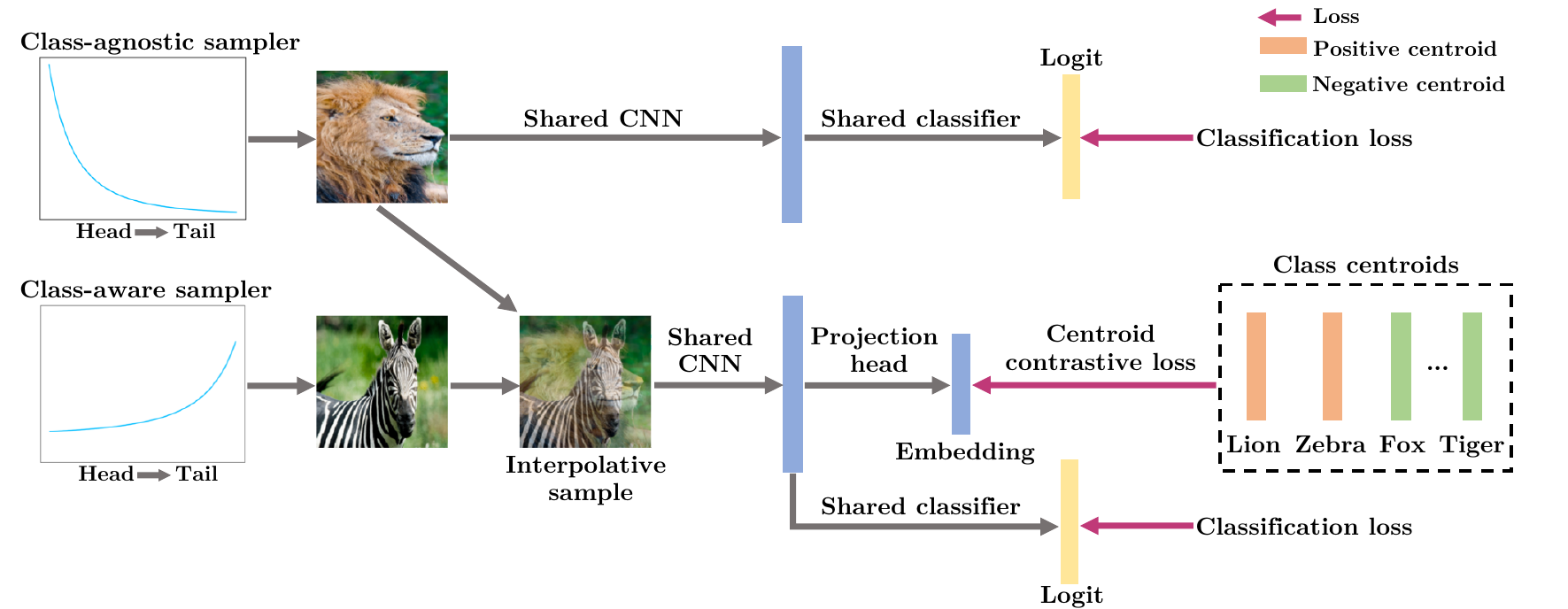}
    \vspace{-3ex}
  \caption
  	{ \small
  		Interpolative centroid contrastive learning framework.
  		The uniform branch (top) focuses more on head-class samples and is trained with the standard cross entropy loss.
  		The interpolative branch (bottom) focuses more on tail-class samples and is trained with an interpolative classification loss and an interpolative centroid contrastive loss.
  		The model parameters are shared between the two branches.
	  } 
  \label{fig:framework}
   \vspace{-2ex}
 \end{figure*} 

%% file: sec_method.tex
\section{Method}
\label{sec:method}
Formally speaking,
for a long-tailed classification task, 
we are given a training dataset $\mathcal{D} = \{ (\Vect{x}_i, y_i)\}_{i=1}^n$, where $\Vect{x}_i$ is an image and $y_i \in \{1, \dots, K\}$ is its class label. The training dataset can be decomposed into $\mathcal{D} = \mathcal{D}^h \cup \mathcal{D}^t$, where $\mathcal{D}^h = \{ (\Vect{x}_i^h, y_i^h)\}_{i=1}^{n^h}$ comprises of head-class samples and $\mathcal{D}^t = \{ (\Vect{x}_i^t, y_i^t)\}_{i=1}^{n^t}$ consists of tail-class samples. Since $n^h \gg n^t$, the model needs to learn strong discriminative representations for tail classes in a low-resource and imbalanced setting, such that it is not overwhelmed by the abundant head-class samples and is able to classify both head and tail classes correctly.

\subsection{Overall framework}
Our proposed long-tailed representation learning framework consists of a uniform and an interpolative branch as illustrated in Figure~\ref{fig:framework}. 
The uniform branch follows the original long-tailed distribution to learn more generalisable representations from data-rich head-class samples,
whereas the interpolative branch focuses more on modelling the tail-class to improve tail-class representation.
Both branches share the same model parameters, which is different from BBN~\cite{bbn}. 
Next, we introduce the components of our framework:
\begin{itemize}[leftmargin=*]
	\setlength\itemsep{0pt} 
	\item A convolutional neural network (CNN) encoder which transforms an image into a feature vector $\Vect{g}_i \in \mathbb{R}^{d_g}$. We experiment with the ResNet~\cite{resnet} model and its variants~\cite{resnext}. The feature $\Vect{g}_i$ is the output from the global average pooling layer.
	\item A projection head which transforms the feature vector $\Vect{g}_i$ into a low-dimensional normalised embedding $\Vect{z}_i \in \mathbb{R}^{d_z}$. Following SimCLR~\cite{simclr}, the projection head is a MLP with one hidden layer of size $d_g$ and ReLU activations. 
	\item A linear classifier with softmax activation which returns a class probability $\Vect{p}_i$ given a feature vector $\Vect{g}_i$.
	\item Class centroids $\Vect{c}^{k} \in \mathbb{R}^{d_z \times K}$ which resides in the low-dimensional embedding space. Similar to MoPro~\cite{li2020mopro}, we compute the centroid of each class as the exponential-moving-average (EMA) of the low-dimensional embeddings for samples from that class. 
	Specifically, the centroid for class $k$ is updated during training by:
	\begin{equation}\label{eq:mom_proto}
		\Vect{c}^{k} \leftarrow m  \cdot \Vect{c}^{k} + (1-m)  \sum_{k=1}^K \mathbbm{1}_{{y_i} = {k}} \cdot \Vect{z}_i,
	\end{equation}
	where $m$ is the momentum coefficient.
	\item A class-agnostic and a class-aware sampler which create interpolative samples.
\end{itemize}

\subsection{Interpolative sample generation}
We utilise two different samplers for interpolative sample generation:
\begin{itemize}[leftmargin=*]
    \item A class-agnostic sampler which selects all samples with an equal probability regardless of the class, thus it returns more head-class samples. We denote a sample returned by the class-agnostic sampler as $(\Vect{x}_i^h, y^h_i)$.
	\item A class-aware sampler which focuses more on tail classes.
	It first samples a class and then select the corresponding samples uniformly with repetition. Let $n^k$ denotes the number of samples in class $k$, the probability $p(k)$ of sampling class $k$ is inversely proportional to $n^k$ as follows:
	\begin{equation}
		p(k) = \frac{({1/n^{k}})^{\gamma}}{\sum_{j=1}^K ({1/n^{j}})^{\gamma}},
	\end{equation} 
	where $\gamma$ is an adjustment parameter. 
	When $\gamma = 0$, the class-aware sampler is equivalent to the balanced sampler in~\cite{shen2016relay}.
	When $\gamma = 1$, it is the reverse sampler in~\cite{bbn}.
    We denote a sample returned by the class-aware sampler as $(\Vect{x}_i^t, y^t_i)$.	
\end{itemize}
An interpolative image $\Vect{x}_i^f$ is formed by linearly combining two images from the class-agnostic and class-aware sampler, respectively. 
\begin{equation}\label{eq:x_fused}
	\Vect{x}^f_i = \lambda \Vect{x}^h_i + (1 - \lambda) \Vect{x}^t_i,  
\end{equation}
where $\lambda \sim \mathcal{U}(0,1)$ is sampled from a uniform distribution.
It is equivalent to the $\mathrm{Beta}(\alpha, \alpha)$ used in Mixup~\cite{mixup} with $\alpha=1$.
Our contrastive learning trains the model such that the representation of the interpolative image is discriminative for both class $y_i^h$ and class $y_i^t$.

\subsection{Interpolative centroid contrastive loss}
Here we introduce the proposed interpolative centroid contrastive loss
which aims to improve long-tailed representation learning.
Given the low-dimensional embedding $\Vect{z}^f_i$ for an interpolative sample $\Vect{x}^f_i$,
we use $\Vect{z}^f_i$ to query the class centroids with contrastive similarity matching.
Specifically,
the probability that the $k$-th class centroid $\Vect{c}^k$ is retrieved is given as:
\begin{equation}
\label{eq:contrast}
p(\Vect{c}^k|\Vect{x}^f_i)=\frac{\operatorname{exp}(\Vect{z}_i^f \boldsymbol{\cdot} \Vect{c}^{{k}}/\tau)}
		{\sum_{j=1}^{K}  \operatorname{exp}(\Vect{z}_i^f \boldsymbol{\cdot} \Vect{c}^j/\tau)},
\end{equation}
where $\tau$ is a scalar temperature parameter to scale the similarity.
Equation~\ref{eq:contrast} can be interpreted as a non-parametric classifier.
Since the centroid is computed as the moving-average of $\Vect{z}_i$,
it does not suffer from the problem of weight imbalance as a parametric classifier does.

Since $\Vect{x}^f_i$ is a linear interpolation of $\Vect{x}^h_i$ and $\Vect{x}^t_i$ (see equation~\ref{eq:x_fused}),
our loss encourages the retrieval of the corresponding centroids of class $y^h_i$ and $y^t_i$.
Thus, the interpolative centroid contrastive loss is defined as:
\begin{equation}\label{eq:cent_fused}
	\mathcal{L}_{cc}^{it} = -\lambda \log(p(\Vect{c}^{y^h_i}|\Vect{x}^f_i)) - (1-\lambda)  \log(p(\Vect{c}^{y^t_i}|\Vect{x}^f_i)).
\end{equation}
The proposed centroid contrastive loss introduces valuable structural information into the embedding space.
The numerator of $p(\Vect{c}|\Vect{x}^f_i)$ reduces the intra-class variance by pulling embeddings with the same class closer to the class centroid.
The denominator of $p(\Vect{c}|\Vect{x}^f_i)$ increases the inter-class variance by pushing an embedding away from other classes' centroids.
Therefore, more discriminative representations can be learned.
 
\input{table/tbl_cifarLT}

\subsection{Classification loss}
Given the classifier's output prediction probability $\Vect{p}(\Vect{x}^h_i)$ for an image $\Vect{x}^h_i$, 
we define the classification loss on the uniform branch as the standard cross entropy loss:
\begin{equation}\label{eq:ce}
	\mathcal{L}_{ce} = 	-\log(p^{y^h_i}(\Vect{x}^h_i)).
\end{equation}
For an interpolative sample $\Vect{x}_i^f$ from the interpolative branch, the classification loss is 
\begin{equation}\label{eq:ce_fused}
	\mathcal{L}_{ce}^{it} = - \lambda \log(p_{i}^{y^h_i}(\Vect{x}_i^f)) - (1-\lambda) \log(p_{i}^{y^t_i}(\Vect{x}_i^f)).
\end{equation}

\subsection{Overall loss} 
During training, 
we jointly minimise the sum of losses on both branches: 
\begin{equation}\label{eq:total_loss}
	\mathcal{L}_{total} = \sum_{i=1}^n \omega_u \mathcal{L}_{ce} +  \omega_{it}(\mathcal{L}^{it}_{ce} + \mathcal{L}^{it}_{cc}),
\end{equation}
where $\omega_{u}$ and $\omega_{it}$ are the weights for the uniform branch and the interpolative branch, respectively.

\subsection{Classifier rebalancing}
\label{subsec:balancing}
Following the decoupled training approach~\cite{decouple-longtail},
we rebalance our classifier after the representation learning stage.
Specifically, we discard the projection head and fine-tune the linear classifier.
The CNN encoder is either fixed or fine-tuned with a smaller learning rate.
In order to rebalance the classifier towards tail classes,
we employ our class-aware sampler.
We denote the sampler's adjustment parameter as $\gamma'$.
Due to more frequent sampling of tail-class samples by the class-aware sampler,
the classifier's logits distribution would shift towards the tail classes at the cost of lower accuracy on head classes.
In order to maintain the head-class accuracy,
we introduce a distillation loss~\cite{hinton2015distilling, tian2019crd} using the classifier from the first stage as the teacher.
The overall loss for classifier rebalancing consists of a cross-entropy classification loss and a KL-divergence distillation loss.
\begin{equation}\label{eq:loss_cb}
		\resizebox{.9\hsize}{!}{$
	\mathcal{L}_{cb} = \sum_{i=1}^n (1-\omega_{d})    \mathcal{L}_{ce}        +         \omega_{d} \tau_{d}^2     \mathcal{L}_{KL} \left(\sigma(\Vect{o}^T/\tau_d), \sigma(\Vect{o}^S/\tau_d)    \right),
	$}
\end{equation}
where $\omega_d$ is the weight of the distillation loss, $\Vect{o}^S$ and $\Vect{o}^T$ are the class logits produced by the student classifier (2nd stage) and the teacher classifier (1st stage) , respectively.
$\tau_d$ is the distillation temperature and $\sigma$ is the softmax function.

For inference, we use a classification network consisting of the CNN encoder followed by the rebalanced classifier.

%% file: table/tbl_cifarLT.tex
\begin{table*}[!t]
	\setlength{\tabcolsep}{10pt} 
	\centering
		\begin{tabular}	{c |c c c | c c c}
			\hline	 
		    Dataset & \multicolumn{3}{c|}{Long-tailed CIFAR100} & \multicolumn{3}{c}{Long-tailed CIFAR10} \\	
			Imbalance ratio & 100 & 50 & 10 & 100 & 50 & 10 \\
			\hline	 
			CE$^{\ast}$ & 38.3 & 43.9 & 55.7 & 70.4 & 74.8 & 86.4  \\  
			Focal Loss$^{\ast}$~\cite{focal} & 38.4 & 44.3 & 55.8 & 70.4 & 76.7 & 86.7 \\ 
			Mixup$^{\ast}$~\cite{mixup} & 39.5 & 45.0 & 58.0 & 73.1 & 77.8 & 87.1 \\
			Manifold Mixup$^{\ast}$~\cite{manifoldmixup}& 38.3 & 43.1 & 56.6& 73.0 & 78.0 & 87.0  \\
			Manifold Mixup (two samplers)$^{\ast}$~\cite{manifoldmixup} & 36.8 & 42.1 & 56.5& 73.1 & 79.2 & 86.8  \\
			CB-Focal$^{\ast}$~\cite{cbloss} & 39.6 & 45.2 & 58.0 & 74.6 & 79.3 & 87.1  \\ 
			CE-DRW$^{\ast}$~\cite{ldam}  & 41.5 & 45.3 & 58.1 & 76.3 & 80.0 & 87.6\\
			CE-DRS$^{\ast}$~\cite{ldam} & 41.6 & 45.5 & 58.1 & 75.6 & 79.8 & 87.4 \\
			LDAM-DRW$^{\ast}$~\cite{ldam} & 42.0 & 46.6 & 58.7  & 77.0 & 81.0 & 88.2 \\
			cRT$^{\dag}$~\cite{decouple-longtail} &     42.3    &    46.8      &   58.1  &   75.7       &   80.4     &   88.3      \\ 
			LWS$^{\dag}$~\cite{decouple-longtail}&   42.3      &     46.4     &   58.1     &    73.0      &   78.5     &    87.7      \\
			BBN~\cite{bbn} & 42.6 & 47.0 & 59.1 & 79.8 & 82.2 & 88.3  \\
			M2m~\cite{kim2020m2m}  & 43.5 & - & 57.6 & 79.1 & - & 87.5 \\
			De-confound-TDE~\cite{tang2020longtailed} & 44.1 & 50.3 & 59.6  & 80.6 & 83.6 & 88.5 \\

			ICCL (ours) & \textbf{46.6} & \textbf{51.6} & \textbf{62.1} &\textbf{82.1} & \textbf{84.7} & \textbf{89.7} \\ 

		\hline	 
	\end{tabular}

    \vspace{-1ex}
	\caption
		{\small	
		Top-1 accuracy on CIFAR100-LT and CIFAR10-LT with different imbalance ratios using ResNet-32 CIFAR. Here, $^{\ast}$ denotes results copied from Zhou \etal ~\cite{bbn}. $^{\dag}$ denotes our reproduced results based on Kang \etal ~\cite{decouple-longtail} setting. 
		}
	\label{tbl:cifarlt}
\vspace{-2ex}
\end{table*}

%% file: sec_experiment.tex
\input{table/tbl_imagenetLT_resneXt50}

\section{Experiments}
\label{sec:experiment}
\subsection{Dataset}

We evaluate our method on three standard benchmark datasets for long-tail recognition as follows:

\textbf{CIFAR-LT.} 
CIFAR10-LT and CIFAR100-LT contain samples from the CIFAR10 and CIFAR100~\cite{cifar} dataset, respectively.
The class sampling frequency follows an exponential distribution.
Following ~\cite{bbn, ldam}, we construct LT datasets with different imbalance ratios of $100$, $50$, and $10$.
Imbalance ratio is defined as the ratio of maximum to the minimum class sampling frequency. The number of training images for CIFAR10-LT with an imbalance ratio of $100$, $50$ and $10$ is $12$k, $14$k and $20$k, respectively. Similarly, CIFAR100-LT has a training set size of $11$k, $13$k and $20$k. Both test sets are balanced with the original size of $10$k.

\textbf{ImageNet-LT.} The training set consists of $1000$ classes with $116$k images sampled from the ImageNet~\cite{ImageNet} dataset.
The class sampling frequency follows a Pareto distribution with a shape parameter of $6$~\cite{oltr}. 
The imbalance ratio is 256.
Despite a smaller training size, it retains ImageNet~\cite{ImageNet} original test set size of $50$k. 

\textbf{iNaturalist 2018.} It is a real-world long-tailed dataset for fine-grained image classification of $8,142$ species~\cite{inat18}. We utilise the official training and test datasets composing of $438$k training and $24$k test images.  

\subsection{Evaluation}
For all datasets, we evaluate our models on the test sets and report the overall top-1 accuracy across all classes. 
To further access the model's accuracy on different classes, we group the classes into splits according to their number of images~\cite{oltr, decouple-longtail}: many ($>100$ images), medium ($20-100$ images) and few ($<20$ images). 

\subsection{Implementation details}
For fair comparison, we follow the same training setup of previous works using SGD optimiser with a momentum of $0.9$. 
For all experiments, we fix class centroid momentum coefficient $m=0.99$, class-aware sampler adjustment parameter $\gamma=0$, $\gamma'=1$, distillation weight $\omega_d=0.5$ and distillation temperature $\tau_d=10$. 
Unless otherwise specified, for the hyperparameters, we set temperature $\tau=0.07$, uniform branch weight $\omega_{u} = 1$, interpolative branch weight $\omega_{it} = 1$, and MLP projection head embedding size $d_z=128$ in the representation learning stage. 
In the classifier balancing stage, we freeze the CNN and fine-tune the classifier using the original learning rate$\times0.1$ with cosine scheduling~\cite{loshchilov2016sgdr} for $10$ epochs.

We also design a warm-up training curriculum.
Specifically,
In the first $T$ epochs, we train only the uniform branch using the cross-entropy loss $\mathcal{L}_{ce}$ and a (non-interpolative) centroid contrastive loss $\mathcal{L}_{cc}=-\log(p(\Vect{c}^{y^h_i}|\Vect{x}^h_i))$.
After $T$ epochs, we activate the interpolative branch and optimise $\mathcal{L}_{total}$ in Equation ~\ref{eq:total_loss}.
The warm-up provides good initialisation for the representations and the centroids.
$T$ is scheduled to be approximately halfway through the total number of epochs. 

\textbf{CIFAR-LT.} We use a ResNet-32~\cite{resnet} as the CNN encoder and follow the training strategies in ~\cite{bbn}. 
We train the model for $200$ epochs with a batch size of $128$. 
The projected embedding size is $d_z=32$. 
We use standard data augmentation which consists of random horizontal flip and cropping with a padding size of $4$. 
The learning rate warms up to $0.1$ within the first $5$ epochs and decays at epoch $120$ and $160$ with a step size of $0.01$. 
We use a weight decay of $2\mathrm{e}{-4}$. 
We set $\tau = 0.3$ and $T$ as $80$ and $100$ epochs for CIFAR100-LT and CIFAR10-LT, respectively.  
$\omega_u$ is set as 0 after warm-up.
In the classifier balancing stage, we fine-tune the CNN encoder using cosine scheduling with an initial learning rate of $0.01$.

\textbf{ImageNet-LT.} We train a ResNeXt-50~\cite{resnext} model for $90$ epochs using a batch size of $256$, a weight decay of $5\mathrm{e}{-4}$,
and a base learning rate of $0.1$ with cosine scheduling. 
Similar to ~\cite{decouple-longtail}, we augment the data using random horizontal flip, cropping and colour jittering. 
We set $T=40$.

\textbf{iNaturalist 2018.} Following ~\cite{decouple-longtail}, we train a ResNet-50 model for $90$ epochs and $200$ epochs using $0.2$ learning rate with cosine decay, $512$ batch size and $1\mathrm{e}{-4}$ weight decay. The data augmentation comprises of only horizontal flip and cropping. $T$ is set as $40$ and $100$ epochs for training epochs of $90$ and $200$, respectively.

\subsection{Results}
\input{table/tbl_inat18_new}
Next we present the results,
where the proposed ICCL achieves significant improved performance on all benchmarks.

\textbf{CIFAR-LT.} 
Table~\ref{tbl:cifarlt} demonstrates that ICCL surpasses existing methods across different imbalance ratios for both CIFAR100-LT and CIFAR10-LT. 
Notably, after the representation learning stage, our approach generally achieves competitive performance compared to existing methods apart from De-confound-TDE~\cite{tang2020longtailed}.  
By balancing the classifier, the performance of ICCL further improves and outperforms De-confound-TDE by $2.5$\% on the more challenging CIFAR100-LT with imbalance ratio of $100$. 

\textbf{ImageNet-LT.} 
Table~\ref{tbl:imagenetlt_resnext} presents the ImageNet-LT results,
where ICCL outperforms the existing methods. 
For ImageNet-LT, we also propose an improved set of hyperparameters which increases the accuracy for existing methods.
Specifically, different from the original hyperparameters used in~\cite{decouple-longtail},
we use a smaller batch size of $256$ and a learning rate of $0.1$.
Furthermore, we find it is better to use original learning rate  $\times0.1$  for classifier balancing.
For fair comparison, we re-implement Decouple methods~\cite{decouple-longtail} and De-confound-TDE~\cite{tang2020longtailed} using 
our settings and obtain better accuracy than those reported in the original papers. 
However, ICCL still achieves the best overall accuracy of $54.0$\% with noticeable 
accuracy gains on medium and few classes.

\textbf{iNaturalist 2018.} 
On the real-world large-scale iNaturalist 2018 dataset, ICCL achieves substantial improvements compared with existing methods as shown in Table~\ref{tbl:inat18_new}.
For $90$ and $200$ epochs, our method surpasses BBN~\cite{bbn} by $4.1$\% and $2.8$\% respectively.
We obtain the split accuracy of BBN based on the checkpoint released by the authors. 
We observe that BBN suffers from a large discrepancy of $21.4$\% between the many and medium class accuracy for $90$ epochs, 
whereas our method has more consistent accuracy across all splits.
Additionally, ICCL obtains a best overall accuracy of $70.5$\% at $90$ epochs which is better than BBN ($69.7$\%) at $180$ epochs.

\subsection{Ablation study}

Here we perform extensive ablation study to examine the effect of each component and hyperparameters of ICCL, and provide analysis on what makes ICCL successful.
\input{table/tbl_abt_component_new}

\input{table/tbl_lambda}

\textbf{Loss components.}
For representation learning, ICCL introduces the interpolative centroid contrastive loss $\mathcal{L}_{cc}^{it}$ and 
the interpolative cross-entropy loss $\mathcal{L}_{ce}^{it}$ as shown in Equation ~\ref{eq:total_loss}. 
In Table~\ref{tbl:abt_component_new},
we evaluate the contribution of each loss components using ImageNet-LT dataset.
We consider many split as the head classes ($>100$ images per class), medium and few splits as the tail classes ($\leq 100$ images per class).
We employ the same classifier balancing technique as described in Section 3.4.
We observe that both $\mathcal{L}_{ce}^{it}$ and $\mathcal{L}_{cc}^{it}$ improve the overall accuracy individually and collectively. 
By comparing with $\mathcal{L}_{ce}^{it}$ only which is equivalent to Mixup~\cite{mixup}, we demonstrate that our loss formulation achieves superior performance.
Additionally, having a warm-up before incorporating interpolative losses provides an extra accuracy boost, especially for the tail classes. 
This aligns with the observation 
in ~\cite{ldam} which suggests that adopting a deferred schedule before re-sampling is better for representation learning.

\input{table/tbl_second_sampler_choice}

\input{table/fig_cls_weight_1st}
\textbf{Interpolation weight $\lambda$.}
In Equation~\ref{eq:x_fused},
We sample the interpolation weight $\lambda\in [0,1]$ from a uniform distribution,
which is equivalent to $\mathrm{Beta}(1, 1)$.
We vary the beta distribution and study its effect on CIFAR100-LT with an imbalance ratio of $100$. 
The resulting accuracy and the corresponding beta distribution are shown in Table~\ref{tbl:lambda}.
Sampling from $\mathrm{Beta}(0.2, 1.0)$ is more likely to return a small $\lambda$, thus the interpolative samples 
contain more information about images from the class-aware sampler.
As we fix $\alpha=\beta$ and increase them from $0.2$ to $2$,
the accuracy increases. 
Good performance can be achieved with $\mathrm{Beta}(1.0, 1.0)$ and $\mathrm{Beta}(2.0, 2.0)$,
where the sampled $\lambda$ is less likely to be an extreme value.

\textbf{Class-aware sampler adjustment parameter $\gamma$}. 
We further investigate the influence of $\gamma$ on representation learning. 
When $\gamma = 0$ and $1$, the class-aware sampler is equivalent to class-balanced sampler~\cite{shen2016relay} 
and reverse sampler~\cite{bbn} respectively.
We include a class-agnostic uniform sampler as the baseline.
Table~\ref{tbl:second_sampler_choice} shows that the interpolative branch sampler should neither focus excessively on the 
tail classes ($\gamma=1$) nor on the head classes (uniform). 
When using either of these two samplers,
the resulting interpolative image might be less informative due to 
excessive repetition of tail-class samples or redundant head-class samples.  

\input{table/tbl_rebalancing_effect}
\input{table/fig_distill_tau}

\textbf{Rebalancing classifier.} 
In Table~\ref{tbl:rebalance_cls_imgnet}, 
we show the effect of classifier rebalancing,
which improves both ICCL and the baseline CE method~\cite{decouple-longtail}.
By learning better tail-class representation, 
ICCL achieves higher overall accuracy compared to~\cite{decouple-longtail} both before and after classifier rebalancing.

\textbf{Classifier balancing parameters.} 
In the classifier balancing stage,
we fix the sampler adjustment parameter $\gamma'=1$, 
and the distillation weight $\omega_d=0.5$.
We study their effects in Table~\ref{tbl:cls_balancing}.
For our ICCL approach, 
using a reverse sampler ($\gamma'=1$) is better 
than a balanced sampler ($\gamma'=0$).
Furthermore, the distillation loss tends to benefit more complex ImageNet-LT and iNaturalist than 
CIFAR-LT datasets.
For the baseline cRT~\cite{decouple-longtail},
applying the reverse sampler and distillation does not give accuracy improvement compared to the default setting (52.4).

\input{table/tbl_cls_balancing}

\textbf{Weight norm visualisation.}
The $L_2$ norms of the weights for the linear classification layer suggest how balanced the classifier is.
Having a high weight norm for a particular class indicates that the classifier is more likely to generate a high logit score for that class.
Figure~\ref{fig:weight_norm1} depicts the weight norm of ICCL and cRT~\cite{decouple-longtail} after the representation learning and classifier balancing stage. 
In both stages, our ICCL classifier has a more balanced weight norm compared with cRT. 
Furthermore, we also plot the norm of our class centroids $\Vect{c}^k$,
which shows that the centroids are intrinsically balanced across different classes.

\textbf{Distillation temperature $\tau_d$.}
In Figure~\ref{fig:distill_tau}, we study how $\tau_d$ affects the accuracy of ICCL on ImageNet-LT. 
We find that the overall accuracy is not sensitive to changes in $\tau_d$.
As $\tau_d$ increases, the teacher's logit distribution becomes more flattened.
Therefore, the accuracy for medium and few class improves,
whereas the accuracy for many class decreases.

%% file: table/tbl_imagenetLT_resneXt50.tex
\begin{table*}[!t]
	
	\centering
	\begin{tabular}	{c|c c c c}
	\hline	 	 	
			Method & Overall & Many & Medium & Few \\
			
		\hline	 
			OLTR$^{\ast}$~\cite{oltr} & 41.9  & 51.0 & 40.8 & 20.8 \\
			Focal Loss$^{\ast}$~\cite{focal}  & 43.7  & 64.3 & 37.1 & 8.2 \\

			NCM~\cite{decouple-longtail}  & 47.3 & 56.6 & 45.3 & 28.1 \\
			$\tau$-norm~\cite{decouple-longtail}  & 49.4 & 59.1 & 46.9 & 30.7 \\ 
			cRT~\cite{decouple-longtail}  & 49.6 & 61.8 & 46.2 & 27.4 \\ 
			LWS~\cite{decouple-longtail} & 49.9 & 60.2 & 47.2 & 30.3 \\
			De-confound-TDE~\cite{tang2020longtailed}  & 51.8 & 62.7 & 48.8 & 31.6 \\ 
	\hline	 
			cRT$^{\dag}$~\cite{decouple-longtail}& 52.4 & 64.3  & 49.1 & 30.7 \\  
			LWS$^{\dag}$~\cite{decouple-longtail}  & 52.5 & 63.0 & 49.6 & 32.8 \\ 
			De-confound-TDE$^{\dag}$~\cite{tang2020longtailed}  & 52.4 &  63.5 &  49.2 & 32.2 \\

			ICCL (ours) & \textbf{54.0}  & 60.7 & \textbf{52.9}  & \textbf{39.0} \\
	\hline	 
	\end{tabular}

    \vspace{-1ex}
	\caption
		{\small	
		Top-1 accuracy on ImageNet-LT using ResNeXt-50. $^{\ast}$ denotes results copied from Tang \etal ~\cite{tang2020longtailed}. $^{\dag}$ denotes our reproduced results using improved settings. 
		The trade-off between head class (i.e. many) and tail class (i.e. medium and few) accuracy is adjustable without affecting the overall accuracy (see Fig.~\ref{fig:distill_tau}).
		}
	\label{tbl:imagenetlt_resnext}
\vspace{-2ex}
\end{table*}

%% file: table/tbl_inat18_new.tex
\begin{table*}[!t]
	
	\centering
	
	\begin{tabular}	{c| c c c  c| c c c c }
		\hline	 	 	
			\multirow{2}{*}{Method} & \multicolumn{4}{c|}{90 Epochs}  & \multicolumn{4}{c}{200 Epochs} \\
			&  Overall & Many & Medium & Few  &  Overall & Many & Medium & Few \\
			\hline	 
			CB-Focal~\cite{cbloss}                                                & 61.1 &  -    &   -   &   -   & - &  -    &   -   &   - \\
			CE-DRS$^{\ast}$~\cite{ldam}                                         & 63.6 &  -    &  -    &   -   & - &  -    &   -   &   - \\
			CE-DRW$^{\ast}$~\cite{ldam}                                    & 63.7 &   -   &  -    &  -    & - &  -    &   -   &   -  \\
			LDAM-DRW~\cite{ldam}                                                & 68.0 &   -   &    -  &  -    & - &  -    &   -   &   - \\
			LDAM-DRW$^{\ast}$~\cite{ldam}                               & 64.6 &   -   &    -  &  -    & 66.1 &   -   &    -  &  -  \\
			NCM~\cite{decouple-longtail}                                          & 58.2 & 55.5 & 57.9 & 59.3  & 63.1 & 61.0 & 63.5 & 63.3 \\
			cRT~\cite{decouple-longtail}                                                  & 65.2 & \textbf{69.0} & 66.0 & 63.2 & 68.2 & \textbf{73.2} & 68.8 & 66.1\\
			$\tau$-norm~\cite{decouple-longtail}                          & 65.6 & 65.6 & 65.3 & 65.9 & 69.3 & 71.1 & 68.9 & 69.3\\
			LWS~\cite{decouple-longtail}                                              & 65.9 & 65.0 & 66.3 & 65.5 & 69.5 & 71.0 & 69.8 & 68.8\\
			BBN~\cite{bbn}                                               & 66.4 & 49.4 &  \textbf{70.8} & 65.3 & 69.7 & 61.7 & \textbf{73.6} & 66.9 \\

			ICCL (ours) &  \textbf{70.5} & 67.6 & 70.2 &  \textbf{71.6} & \textbf{72.5} & 72.1 & 72.3 & \textbf{72.9} \\

		\hline	 
	\end{tabular}
    \vspace{-1ex}
	\caption
		{\small	
		Top-1 accuracy on iNaturalist 2018 using ResNet-50 for 90 epochs and 200 epochs. $^{\ast}$ denotes results copied from Zhou \etal ~\cite{bbn}
		which uses 90 and 180 epochs.
        The trade-off between head class (i.e. many) and tail class (i.e. medium and few) is adjustable without affecting the overall accuracy (see Fig.~\ref{fig:distill_tau}).
		}
	\label{tbl:inat18_new}
\vspace{-2ex}
\end{table*}		

%% file: table/tbl_abt_component_new.tex
\begin{table}[!t]
\small

	\centering

	\begin{tabular}	{ c c c c |c c c }
	\hline	 	 
		    $\mathcal{L}_{ce} $ & $\mathcal{L}^{it}_{ce}$ & $\mathcal{L}^{it}_{cc}$&     Warm-up   &   Overall  & Head & Tail \\
	\hline	 
		 	\checkmark		& 				& 						& 		 	& 51.3       & 60.6 & 45.5\\
		 									& 		\checkmark			& 					&	& 51.6 & 58.3	 & 47.4 \\
								& 		\checkmark			& 					&	\checkmark		& 	51.7 & 57.9 &  47.9\\
			\checkmark	& 		 	& 		\checkmark			&\checkmark   	&   52.4  	       & 59.9 &  47.7\\
			\checkmark	& 		\checkmark		& 				& 	\checkmark	&  53.4             & 61.1 & 48.6\\
			\checkmark	& 		\checkmark		& 		\checkmark		&	  	&  53.6 & 61.2 & 48.8\\
			 \checkmark	& 		\checkmark		& 		\checkmark		&	\checkmark   	& 54.0    & 60.7 & 49.8\\
	\hline	 
	\end{tabular}

	\vspace{-1ex}
	\caption
		{\small	
		Ablation study on different components of ICCL on ImageNet-LT.
		Head denotes the many split, whereas tail includes the medium and few splits.
		The proposed $\mathcal{L}^{it}_{ce}$, $\mathcal{L}^{it}_{cc}$, and warm-up all contribute to accuracy improvement.
		Using only $\mathcal{L}^{it}_{ce}$ is equivalent to Mixup~\cite{mixup}. 
		}
	\label{tbl:abt_component_new}
\vspace{-2ex}
\end{table}

%% file: table/tbl_lambda.tex
\begin{table}[!t]
	\begin{minipage}[b]{0.5\linewidth}
		\centering
			\begin{tabular}	{c |c }
			\hline	 	 
			$\mathrm{Beta}(\alpha, \beta)$  &    CIFAR100-LT \\
		\hline	 
			(0.2, 1.0)			& 43.6 \\
			(0.2, 0.2)			& 43.8\\
			(0.6,0.6) 		&   45.4\\
			(1.0,1.0)		&     46.6\\
			(2.0, 2.0) 		&   46.8\\
		\hline	 
			\end{tabular}

	\end{minipage}\hfill
	\begin{minipage}[!b]{0.44\linewidth}
		\centering
		\includegraphics[width=38mm]{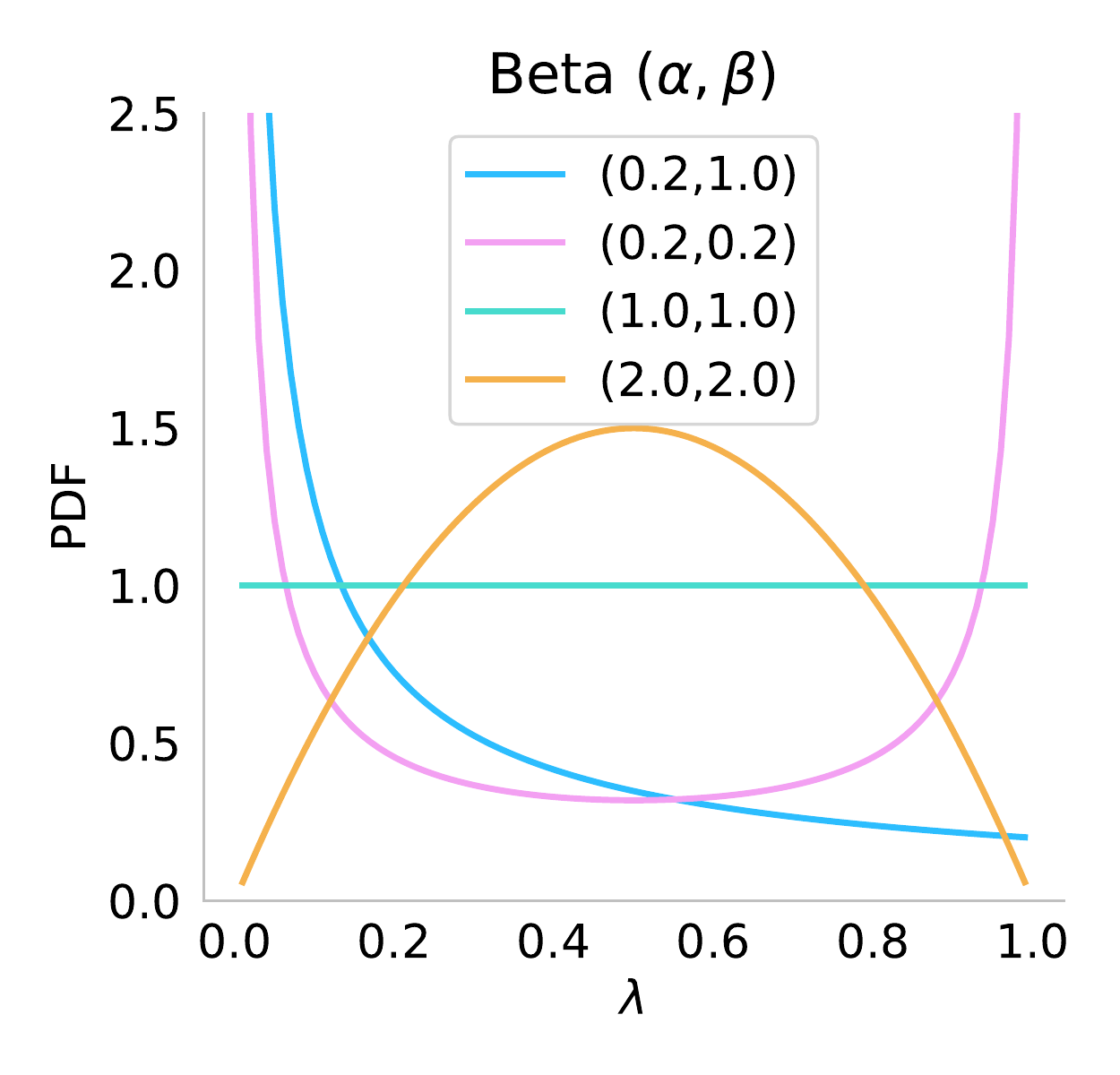}

	\end{minipage}
	\vspace{-2.3ex}
	\caption
		{\small	
		Effect of sampling $\lambda$ from different $\mathrm{Beta}(\alpha, \beta)$ on CIFAR100-LT.
		$\lambda$ determines the weighting of the two samples for a given interpolative sample.  
		}
    \label{tbl:lambda}
\vspace{-2.5ex}
\end{table}

%% file: table/tbl_second_sampler_choice.tex
\begin{table}[!t]
		\setlength\tabcolsep{4pt}
	\centering
	\resizebox{1\columnwidth}{!}{  
	\begin{tabular}	{c |c |c |c |c }
	\hline	 	 
		  Sampler &     CIFAR100-LT    & CIFAR10-LT & ImageNet-LT & iNaturalist \\
\hline	 
			Uniform & 44.7 & 79.9 & 52.8& 69.4 \\

			$\gamma=0$ & 46.6 & 82.1 & 54.0 & 70.5 \\
			$\gamma=0.5$ & 46.2 & 81.6 & 54.1 & 70.2\\
			$\gamma=1.0$ & 46.2 & 81.1 & 53.1 & 70.1 \\
	\hline	 
	\end{tabular}
}
    \vspace{-1ex}
	\caption
		{\small	
		Adjustment parameter $\gamma$ of the interpolative branch class-aware sampler.
		Focus excessively on head-class (uniform) or tail-class samples ($\gamma=1$)
		leads to worse performance.
		}
	\label{tbl:second_sampler_choice}
\vspace{-2.5ex}
\end{table}

%% file: table/fig_cls_weight_1st.tex
\begin{figure*}[!t]

	\centering
	\includegraphics[width=0.8\textwidth]{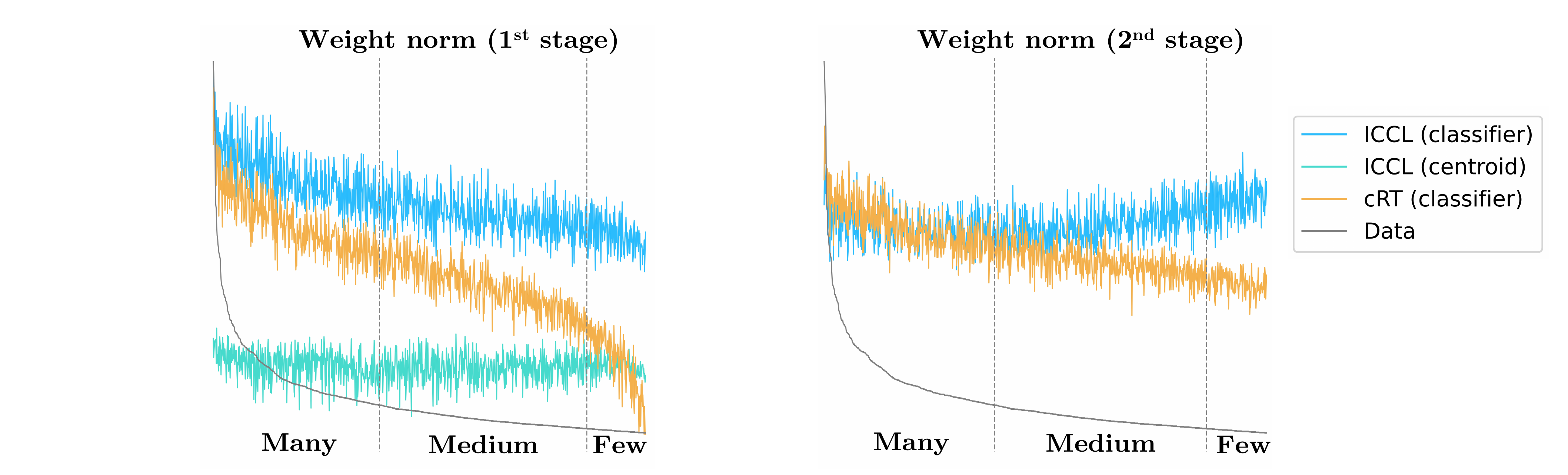}
    \vspace{-2ex}
  \caption
  	{ \small
	Visualisation of classifer's weight norm and centroid's norm of ICCL after the representation learning (left) and classifier balancing stage (right) on ImageNet-LT. Comparing with cRT~\cite{decouple-longtail}, the weight norm of our ICCL classifier is more balanced. Additionally, the class centroids have intrinsically balanced norm.
	  } 
  \label{fig:weight_norm1}
   \vspace{-2.5ex}
 \end{figure*} 

%% file: table/tbl_rebalancing_effect.tex
\begin{table}[!t]
	
	\centering
	\resizebox{1\columnwidth}{!}{
	\begin{tabular}	{c|c| c c c c}
		\hline	 	 	
			Method & Rebalancing & Overall & Many & Medium & Few \\
		\hline	 
			CE~\cite{decouple-longtail} & - &  46.7 & 68.1 & 40.2 & 9.0 \\
			ICCL (ours)  & -	& \textbf{50.5} & \textbf{68.5} & \textbf{44.4} & \textbf{20.8}  \\
		\hline	 
			
			CE~\cite{decouple-longtail}& \checkmark& 52.4 & \textbf{64.3}  & 49.1 & 30.7 \\ 
			ICCL (ours) & \checkmark	& \textbf{54.0}  & 60.7 & \textbf{52.9}  & \textbf{39.0} \\
	\hline	 
	\end{tabular}
}
    \vspace{-1ex}
	\caption
		{\small	
		Effect of classifier rebalancing on ImageNet-LT. ICCL learns better tail-class representation which leads to higher tail-class (i.e. medium and few) accuracy after classifier rebalancing.
		}
	\label{tbl:rebalance_cls_imgnet}
\vspace{-2ex}
\end{table}	

%% file: table/fig_distill_tau.tex
\begin{figure}[!t]

	\centering
	\includegraphics[width=0.45\textwidth]{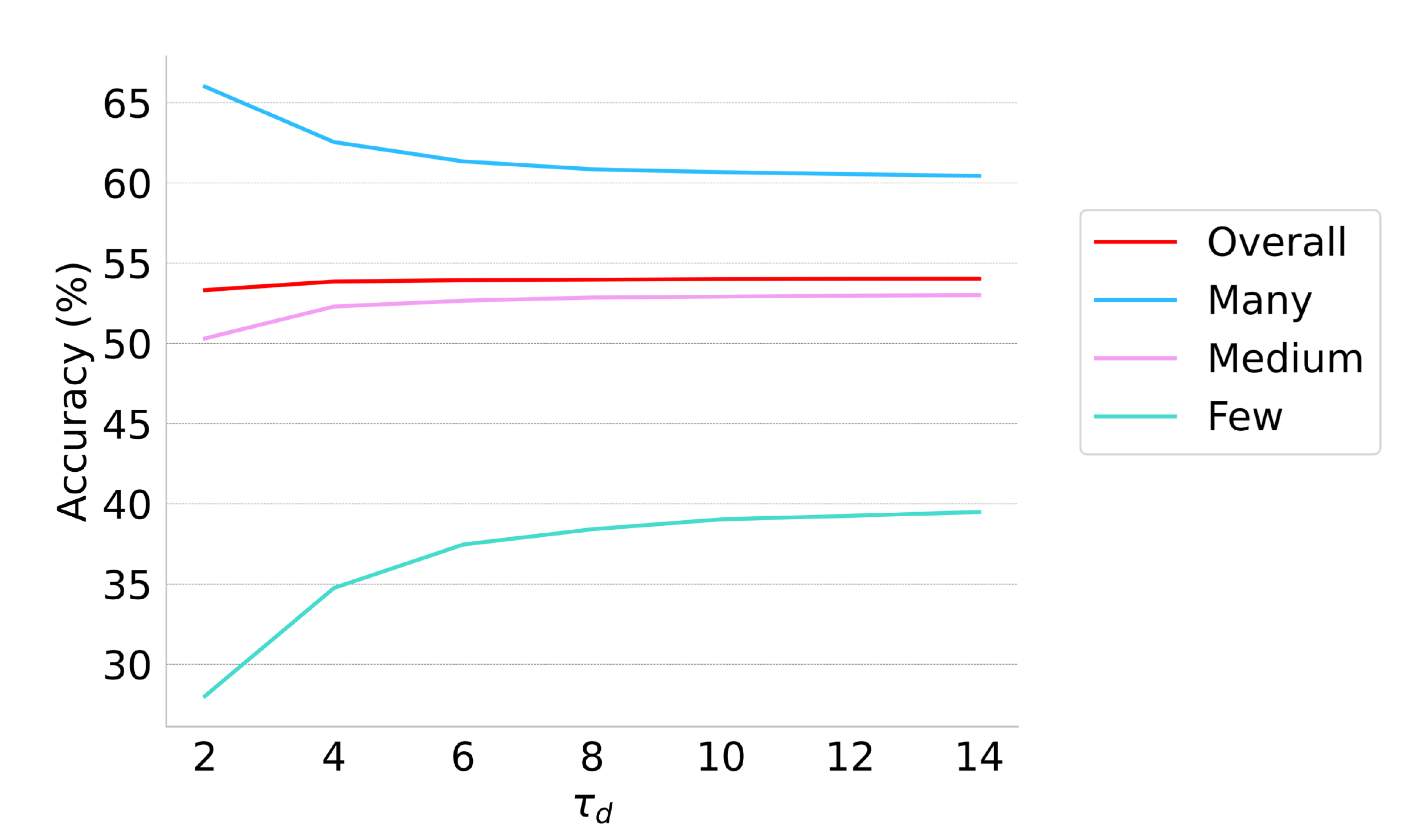}
	\vspace{-1ex}

	\caption
	{ \small
		Effect of distillation temperature $\tau_d$ on ImageNet-LT. ICCL's overall accuracy is not sensitive to $\tau_d$ variation.
	} 
	\label{fig:distill_tau}
	\vspace{-3ex}
\end{figure} 

%% file: table/tbl_cls_balancing.tex
\begin{table}[!t]
		\setlength\tabcolsep{4pt}
	\centering
	\resizebox{1\columnwidth}{!}{  
	\begin{tabular}	{ c c |c |c |c |c c}
		\hline	 	 
		    \multirow{2}{*}{$\gamma'$} &  \multirow{2}{*}{$\omega_d$} &    CIFAR100-LT  & CIFAR10-LT  & iNaturalist  & \multicolumn{2}{c}{ImageNet-LT} \\
			\cmidrule{3-7}
		   &  & ICCL & ICCL & ICCL & ICCL & cRT~\cite{decouple-longtail} \\
			\midrule
			0 & 0			& 45.3 & 77.5 & 69.5 & 53.7 & 52.4\\
			0 & 0.5  & 45.0 &  77.6 &  69.5  & 53.2 &  52.2 \\
			1 & 0		&  47.1 &  82.3 &  70.2 & 53.6 & 49.6 \\
			1 & 0.5  & 46.6 &82.1  & 70.5 & 54.0 & 51.3 \\
	\hline	 
	\end{tabular}
}
\vspace{-1ex}
	\caption
		{\small	
		Ablation study for classifier balancing methods.
		ICCL benefits from using a reverse sampler ($\gamma'=1$) 
		and knowledge distillation ($\omega_d=0.5$), especially for 
		the more complex ImageNet-LT and iNaturalist datasets.
		}
	\label{tbl:cls_balancing}
\vspace{-2.5ex}
\end{table}

%% file: sec_conclusion.tex
\section{Conclusion}
\label{sec:conclusion}
In this work, we propose an interpolative centroid contrastive learning technique for long-tailed representation learning.
By utilising class centroids and interpolative losses,
we strengthen the discriminative power of the learned representations, leading to improved classification accuracy.
We demonstrate the effectiveness of our approach
with significant improvement on multiple long-tailed classification benchmarks.

%% file: sec_acknowledgment.tex
\section{Acknowledgments}

Anthony Meng Huat Tiong is supported by Salesforce and the Singapore Economic Development Board under the Industrial Postgraduate Programme. Boyang Li is supported by the Nanyang Associate Professorship and the National Research Foundation Fellowship (NRF-NRFF13-2021-0006), Singapore. Any opinions, findings, conclusions, or recommendations expressed in this material are those of the authors and do not reflect the views of the funding agencies.